\documentclass{article} %
\usepackage{colm2024_conference}

\usepackage{booktabs}
\usepackage{graphicx}
\usepackage{enumitem}
\usepackage{wrapfig}
\usepackage{algorithm}
\usepackage{algpseudocode}
\usepackage{wrapfig}
\usepackage{float}
\usepackage{microtype}
\usepackage{amsmath}
\usepackage{amssymb}
\usepackage{colortbl}
\usepackage[utf8]{inputenc}
\definecolor{lightgray}{rgb}{0.9,0.9,0.9}
\usepackage{caption}
\usepackage{subcaption}
\usepackage{xcolor}
\usepackage{setspace}
\usepackage{url}
\usepackage{multirow}
\usepackage{colortbl}
\usepackage{tabularx}
\usepackage{blindtext}
\usepackage{pgfplots}
\pgfplotsset{compat=1.18} 
\usepackage{tikz}
\usetikzlibrary{er,positioning,bayesnet}
\usepackage{makecell}
\usepackage{tipa}
\usepackage{siunitx}
\usepackage{nicefrac}
\usepackage{tocloft}
\usepackage{listings}
\usepackage[raster,skins]{tcolorbox} %
\usepackage{xltabular}
\usepackage{adjustbox}
\usepackage{xurl}

\usepackage{amsmath,amsfonts,bm}

\def\figref#1{figure~\ref{#1}}

\def\eqref#1{equation~\ref{#1}}

\def\1{\bm{1}}

\DeclareMathAlphabet{\mathsfit}{\encodingdefault}{\sfdefault}{m}{sl}
\SetMathAlphabet{\mathsfit}{bold}{\encodingdefault}{\sfdefault}{bx}{n}

\makeatletter
\DeclareRobustCommand\onedot{\futurelet\@let@token\@onedot}
\def\@onedot{\ifx\@let@token.\else.\null\fi\xspace}

\def\etc{\emph{etc}\onedot}

\makeatother

\title{Hunyuan3D 2.5: Towards High-Fidelity \\ 3D Assets Generation with  Ultimate Details}

\author{
\bf Tencent Hunyuan3D 
}

\newcommand{\shortname}{Hunyuan3D 2.5\xspace}

\begin{document}

\maketitle

\begin{abstract}
In this report, we present \shortname, a robust suite of 3D diffusion models aimed at generating high-fidelity and detailed textured 3D assets. \shortname follows two-stages pipeline of its previous version Hunyuan3D 2.0, while demonstrating substantial advancements in both shape and texture generation.
In terms of shape generation, we introduce a new shape foundation model -- LATTICE, which is trained with scaled high-quality datasets, model-size, and compute. Our largest model reaches 10B parameters and generates sharp and detailed 3D shape with precise image-3D following while keeping mesh surface clean and smooth, significantly closing the gap between generated and handcrafted 3D shapes.
In terms of texture generation, it is upgraded with phyiscal-based rendering (PBR) via a novel multi-view architecture extended from Hunyuan3D 2.0 Paint model.
Our extensive evaluation shows that \shortname significantly outperforms previous methods in both shape and end-to-end texture generation.
\end{abstract}

\begin{figure}[h]
\centering
\includegraphics[width=0.83\textwidth]{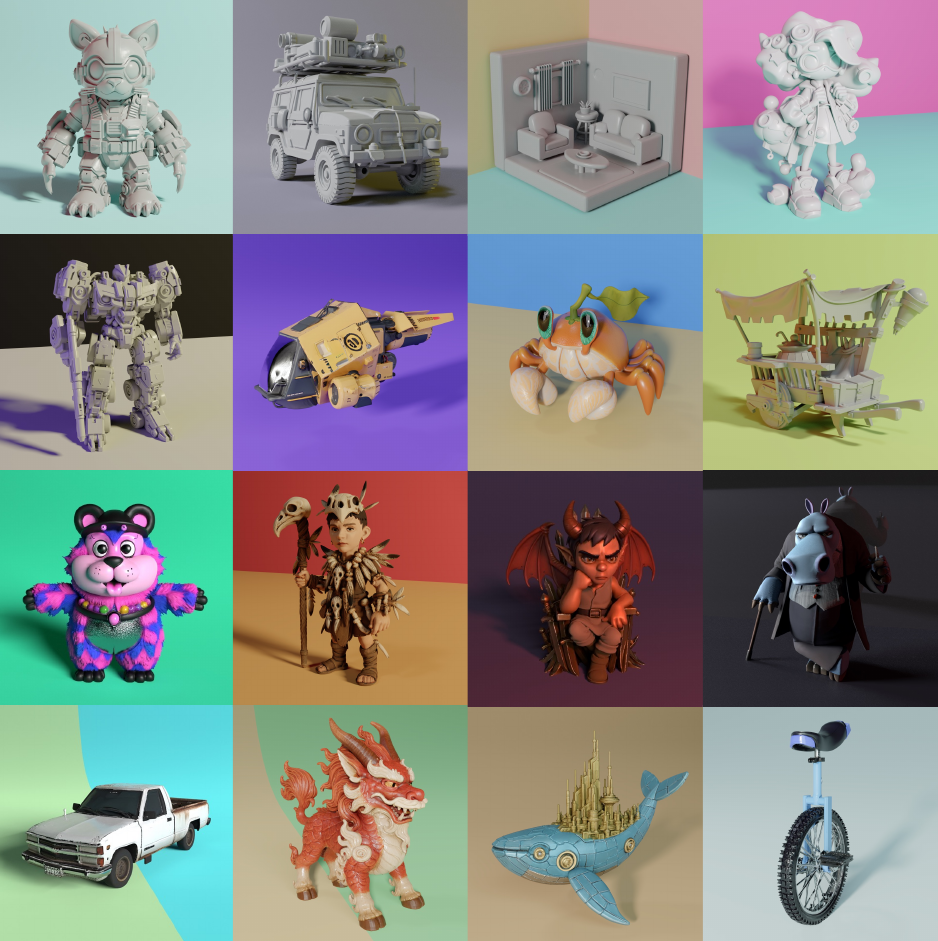}
\caption{High quality 3D assets generated by \shortname.}
\end{figure}

\clearpage

\section{Introduction}

3D generation has rapidly developed in recent years, becoming a core driver of innovation and growth across various industries. From game development to embodied AI, from film special effects to virtual reality, the application scenarios of 3D models continue to expand, demonstrating their immense potential and value. With advancements in artificial intelligence, 3D generation has become more efficient and powerful, particularly in areas such as automated modeling and texturing, further simplifying the creative process and enhancing production efficiency.

\begin{wrapfigure}{r}{0.5\textwidth}
    \centering
    \includegraphics[width=0.5\textwidth]{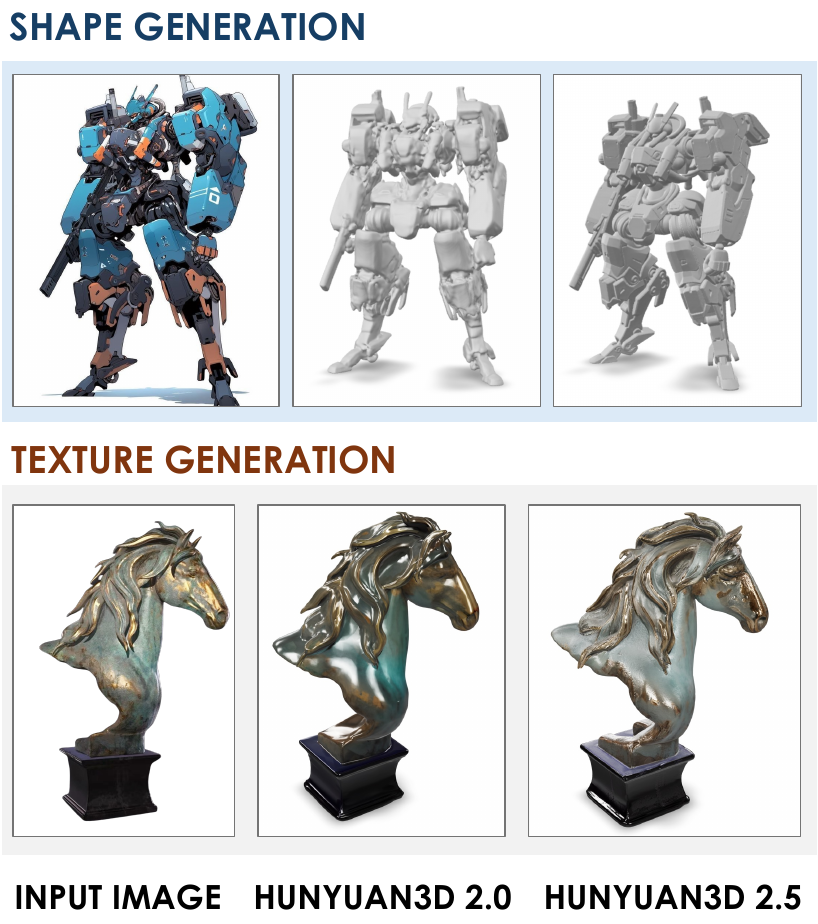}
    \caption{Drawbacks of existing methods: failure at detail generation and incorrect PBR.}
    \label{fig:drawbacks}
\end{wrapfigure}

Notably, recent 3D shape diffusion models based on 3dshape2vecset~\citep{zhang20233dshape2vecset} have pioneered a revolution in the 3D shape generation pipeline, as demonstrated by works such as  CLAY~\citep{zhang2024clay}, Hunyuan3D 2.0~\citep{zhao2025hunyuan3d}, and TripoSG~\citep{li2025triposg}. 
Direct3D~\citep{wu2024direct3d},from another aspect, shown the potential of compressing and generating the shape via triplane.
More recently, Trellis~\citep{xiang2024structured} has emerged as a promising pipeline for high-quality textured 3D generation, leveraging its invented structured 3D latents as representations. Nevertheless, existing models are still limited for generating complex objects with finegrained details as demonstrated in \figref{fig:drawbacks}. It remains an open problem how we could generate high-fidelity and detailed shape while maintaining smooth surface and sharp edges.  

High-quality textures play a crucial role in enhancing the visual realism and detail representation of 3D assets. Recently, Numerous texture generation methods based on multiview diffusion~\citep{zhao2025hunyuan3d,huang2024mvadapter,vainer2024jointlygeneratingmultiviewconsistent, li2024era3d, tang2025mvdiffusion++, long2024wonder3d, wang2023imagedream, shimvdream, shi2023zero123++} have emerged, alleviating global consistency issue of inpainting-based methods~\citep{huang2024material,wu2024texro,zhang2024mapa,ceylan2024matatlas,zeng2024paint3d,chen2023text2tex,richardson2023texture} and synchronization techniques~\citep{liu2025vcd,gao2024genesistex,liu2024text,zhang2024texpainter}. However, challenges remain in generating highly consistent multiview images, which can lead to artifacts and seams during the fusion and baking stages. Moreover, traditional RGB textures can no longer meet the demands for photorealistic 3D asset generation, while PBR material generation solution is not available in open source community.

This report presents \textbf{\shortname}, a robust suite of 3D diffusion models aimed at generating high-fidelity and detailed textured 3D assets. 
\shortname builds upon the two-stage pipeline of its predecessor Hunyuan3D 2.0~\citep{zhao2025hunyuan3d} and 2.1~\citep{hunyuan3d2025hunyuan3d21imageshighfidelity}, while showcasing significant advancements in both shape generation and texture synthesis.

In the first stage of shape generation, we introduce a new shape foundation model -- LATTICE, which has been trained on large-scale, high-quality datasets with increased model size and computational resources. We found that this new model exhibits stable improvement when scaling up the model. Benifit from these characteristics, our largest model generates detailed and sharp 3D shape with precise alignment to corresponding images while maintaining clean and smooth surfaces, significantly closing the gap between generated and handcrafted 3D shapes.

In the second stage of texture generation, we extend the Hunyuan3D 2.0~\citep{zhao2025hunyuan3d} and 2.1~\citep{hunyuan3d2025hunyuan3d21imageshighfidelity} texture generation model into a high-fidelity material generation framework. Adhering to the principled BRDF model, our approach produces multi-view albedo, roughness, and metallic maps simultaneously. This aims to precisely describe the surface reflection properties of generated 3D assets and accurately simulate geometric microsurface distributions, thereby achieving more realistic and detailed rendering results. Furthermore, we introduce a dual-phase resolution enhancement strategy to strengthen texture-geometry coordination, thereby improving end-to-end visual quality.

We perform extensive quantitative and qualitative evaluations as well as user studies in terms of shape generation and end-to-end texture generation, across diverse range of in-the-wild input images. The results demonstrate that \shortname outperforms state-of-the-art open-source and closed-source commercial models. 

\section{Method}
\begin{figure}[t]
    \centering
    \includegraphics[width=1\textwidth]{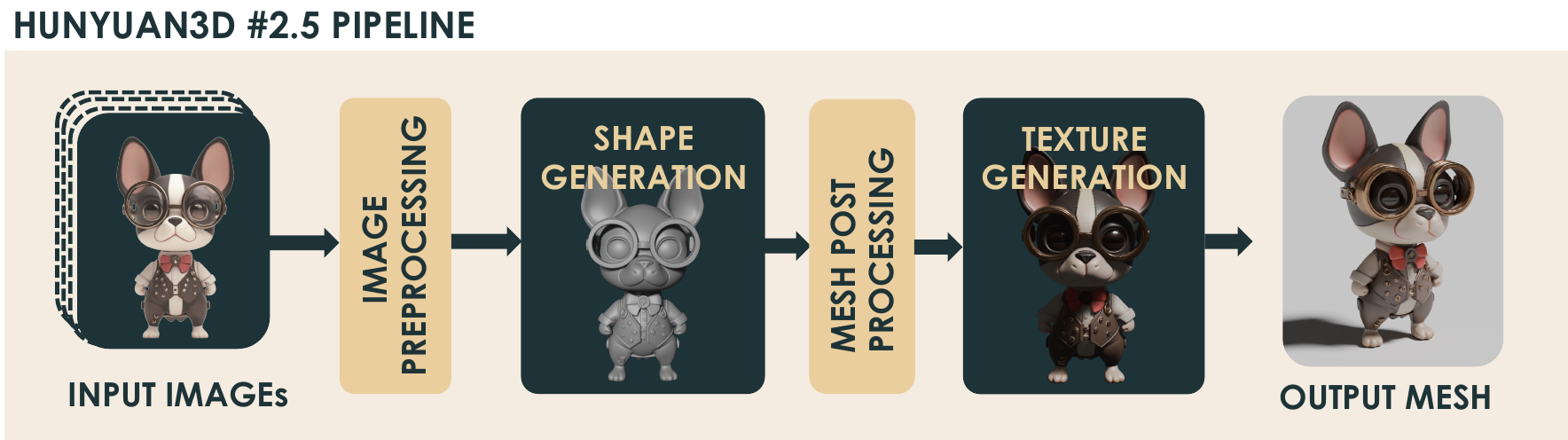}
    \caption{\textbf{Overview of \shortname pipeline}. It separates the 3D asset generation into two stages: first, it generates the shape, and then it creates the texture based on that shape.}
    \label{fig:pipeline}
\end{figure}

\begin{wrapfigure}{r}{0.5\textwidth}
    \vspace{-10mm}
    \centering
    \includegraphics[width=0.5\textwidth]{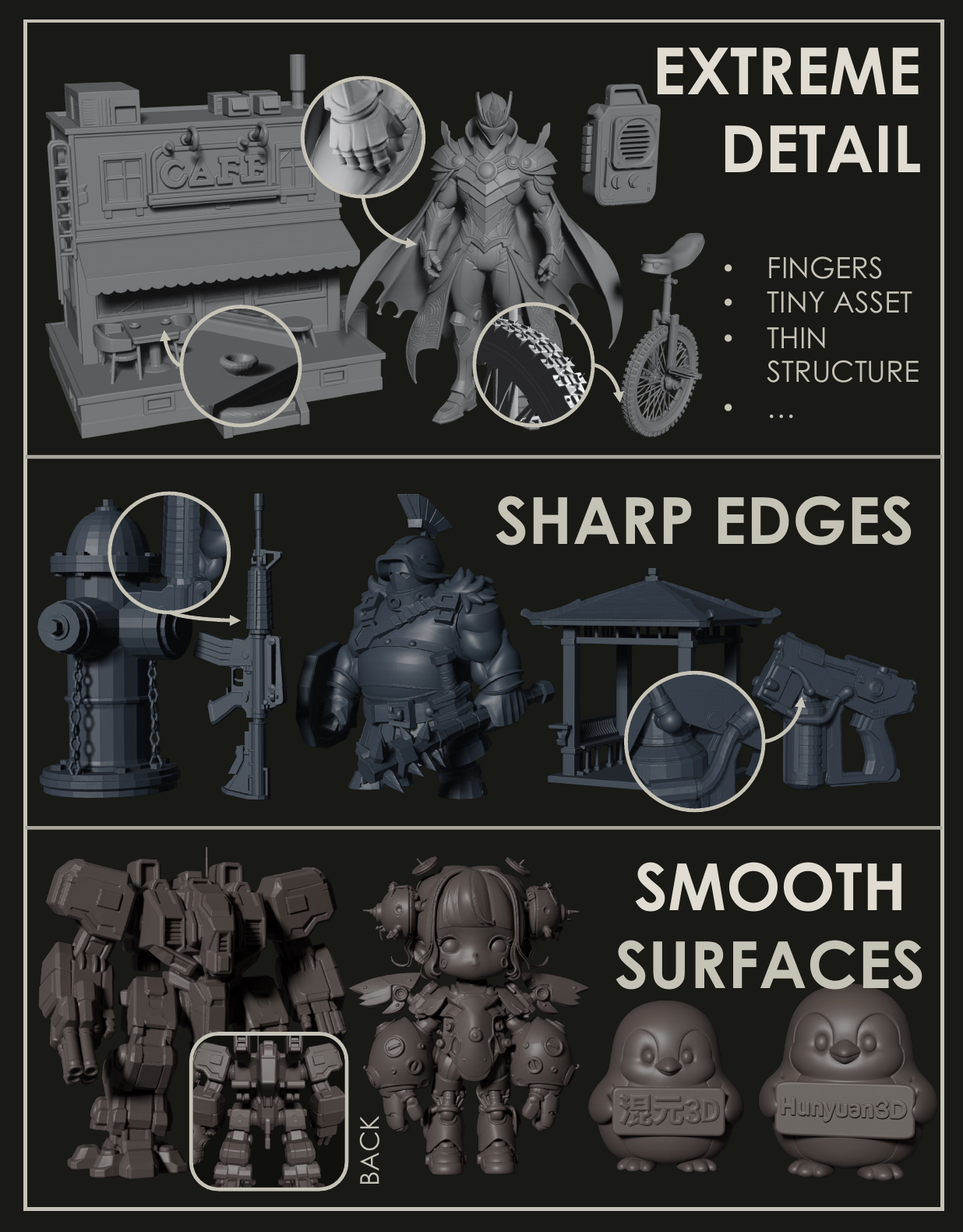}
    \caption{Illustration of major features of the new shape generation model in \shortname.}
    \label{fig:shape_features}
    \vspace{-10mm}
\end{wrapfigure}

\shortname is an image-to-3D generation model, which follows the same overall architecture of Hunyuan3D 2.0~\citep{zhao2025hunyuan3d}, as shown in \figref{fig:pipeline}. In a nutshell, the input image is first processed by an image processor to remove the background and perform proper resizing. Then, a shape generation model is conditioned on the input image and generates the 3D mesh without texture. The mesh is further processed to extract normal, UV map, and \etc. After that, a texture generation model is called to generate the texture with previous outputs.

\subsection{Detailed Shape Generation}

\shortname introduces a new shape generation model -- \href{https://github.com/Zeqiang-Lai/LATTICE}{LATTICE}, which is a large-scale diffusion model capable of producing high-fidelity, detailed shapes with sharp edges and smooth surfaces from either a single image or four multi-view images. Trained on an extensive and high-quality 3D dataset featuring complex objects, the model is designed to generate exceptional detail. To ensure efficiency, we employ guidance and step distillation techniques to reduce inference time.

\textbf{Extreme Detail.} Benfit from scaling up, \shortname can generate fine-grained details at an unprecedented level. In the first row of \figref{fig:shape_features}, we present several examples generated by our model. As shown, the model achieves a level of accuracy approaching that of handcrafted designs, such as the correct number of fingers, the detailed bicycle wheel pattern, and even a bowl within a large scene.

\textbf{Smooth Surfaces \& Shape Edges.} Existing models~\citep{zhao2025hunyuan3d, li2025triposg, xiang2024structured} often struggle to generate sharp edges while maintaining smooth, clean surfaces, particularly for complex objects. In contrast, \shortname strikes an excellent balance, as demonstrated in the second and third rows of \figref{fig:shape_features}.

\subsection{Realistic Texture Generation}
\begin{figure}[t]
    \centering
    \includegraphics[width=1\textwidth]{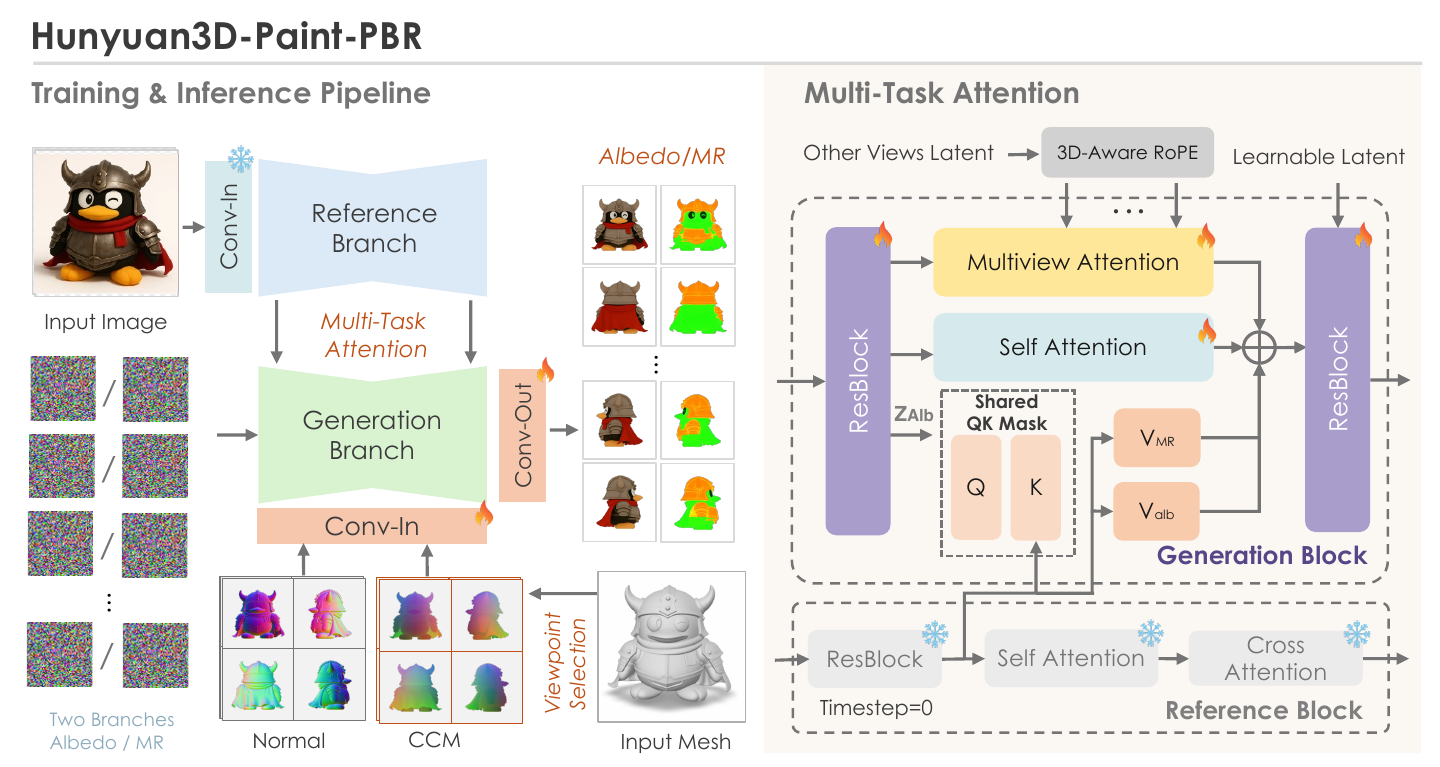}
    \caption{Overview of material generation framework. }
    \label{fig:texture_pipe}
\end{figure}

We propose a novel material generation framework in Hunyuan3D 2.5, which is extended on the foundation of multiview PBR texture generation architecture of hunyuan3D 2.1~\citep{hunyuan3d2025hunyuan3d21imageshighfidelity}. As shown in \figref{fig:texture_pipe}, our model takes normal map and CCM rendered by 3D mesh as geometry conditions, and a reference image as guidance, generating high-quality PBR material maps as texture. We inherit 3D-aware RoPE in~\citep{feng2025romantexdecoupling3dawarerotary} to enhance cross-view consistency for seamless texture map generation.

\paragraph{Multi-Channel Material Generation.}
We introduce learnable embeddings for three material maps: albedo, MR, of which the MR channel is the combation expression of metallic and roughness. Specifically, three independent embeddings $\mathbf{E}_{\text{albedo}}$, $\mathbf{E}_{\text{mr}}$, and $\mathbf{E}_{\text{normal}} \in \mathbb{R}^{16 \times 1024}$ are initialized and subsequently injected into the respective channels via cross-attention layers. The embedding and attention modules are trainable, allowing the network to effectively model the distribution of three materials separately.

Although material channels exhibit significant domain gaps, maintaining spatial correspondence is crucial across different levels, from semantic to pixel-level alignment. To address this challenge, we propose a dual-channel attention mechanism that ensures spatial alignment among generated albedo and metallic-roughness (MR).

After systematically examining the reference attention module, we found that the main cause of multi-channel misalignment lies in the misaligned attention masks. Therefore, we intentionally share the attention mask among multiple channels while varying the value computation in the output calculation.
Specifically, since the basecolor branch contains the most semantically similar information to the reference image (both exist in the common RGB color space), we utilize the attention mask calculated from the basecolor channel and apply it to guide the reference attention in the other two branches, as formulated below:

\begin{align}
    \text{M}_{attn} = &\text{Softmax}\left(\frac{Q_{albedo}K_{ref}^T}{\sqrt{d}}\right)\\
    z_{albedo}^{new} = z_{albedo} &+ \text{MLP}_{albedo}\left[\text{M}_{attn}\cdot V_{albedo}\right], \notag \\
    z_{MR}^{new} = z_{MR} &+ \text{MLP}_{MR}\left[\text{M}_{attn}\cdot V_{MR}\right]
    \label{eq:mcaa_albedo}
\end{align}
This design enables the generated albedo and MR features to maintain spatial coherence while being guided by the reference image's information.
Building upon this framework, we incorporated an illumination-invariant consistency loss during training to enforce the disentanglement of material properties and illumination components. For more information, please refer to ~\citep{he2025materialmvpilluminationinvariantmaterialgeneration}.

\paragraph{Geometric Alignment.} 
The alignment of textures with geometry critically impacts the visual integrity and aesthetic quality of 3D assets. However, achieving precise texture-geometry alignment presents considerable challenges, particularly for complex, high-polygon geometry. A key observation from our analysis is that higher-resolution images preserve richer high-frequency geometric details while mitigating VAE compression losses, thereby significantly enhancing geometric conditioning. 
Nevertheless, training with high-resolution multi-view images demands substantial memory resources, which necessitates reducing the number of views during training and consequently deteriorates the model's capability for dense-view inference.

To address this challenge, we propose a dual-phase resolution enhancement strategy that progressively improves texture-geometry alignment quality while maintaining computational feasibility.
In the first phase, we employ a conventional multi-view training approach using 6-view 512×512 images, following the methodology of Hunyuan3D-2.0~\cite{zhao2025hunyuan3d}. This phase establishes a solid foundation for multi-view consistency and basic texture-geometry correspondence.

In the second phase, we implement a zoom-in training strategy that enables the model to capture high-quality details while preserving the multi-view training benefits from the first phase. Specifically, we randomly zoom into both the reference image and multi-view generated images during training. This approach allows the model to learn fine-grained texture details without requiring full high-resolution training from scratch, thereby circumventing the memory constraints associated with direct high-resolution multi-view training.

During inference, we leverage multi-view images at up to 768×768 resolution, accelerated by the UniPC sampler~\citep{zhao2023unipc} for efficient high-quality generation.

\section{Evaluation}

To comprehensively assess the performance of \shortname, we performed evaluations from two key perspectives: (1) 3D shape generation, and (2) textured 3D asset generation.

\subsection{Shape Generation}


\textbf{Competing Methods.} 
We compare with open-source baselines,  Michelangelo~\citep{zhao2024michelangelo}, Craftsman 1.5~\citep{li2024craftsman}, Trellis~\citep{xiang2024structured}, and Hunyuan3D-2~\citep{zhao2025hunyuan3d}, and closed-source baselines are Commerical Model 1, and Commerical Model 2.

\textbf{Metrics.}
To assess the performance of shape generation, we utilize ULIP~\citep{xue2023ulip} and Uni3D~\citep{zhou2023uni3d} to calculate the similarity between the generated mesh and the input images (ULIP-I and Uni3D-I) as well as image prompts synthesized by the vision-language model~\citep{chen2024internvl} (ULIP-T and Uni3D-T).

\begin{figure}[H]
    \centering
    \includegraphics[width=1\textwidth]{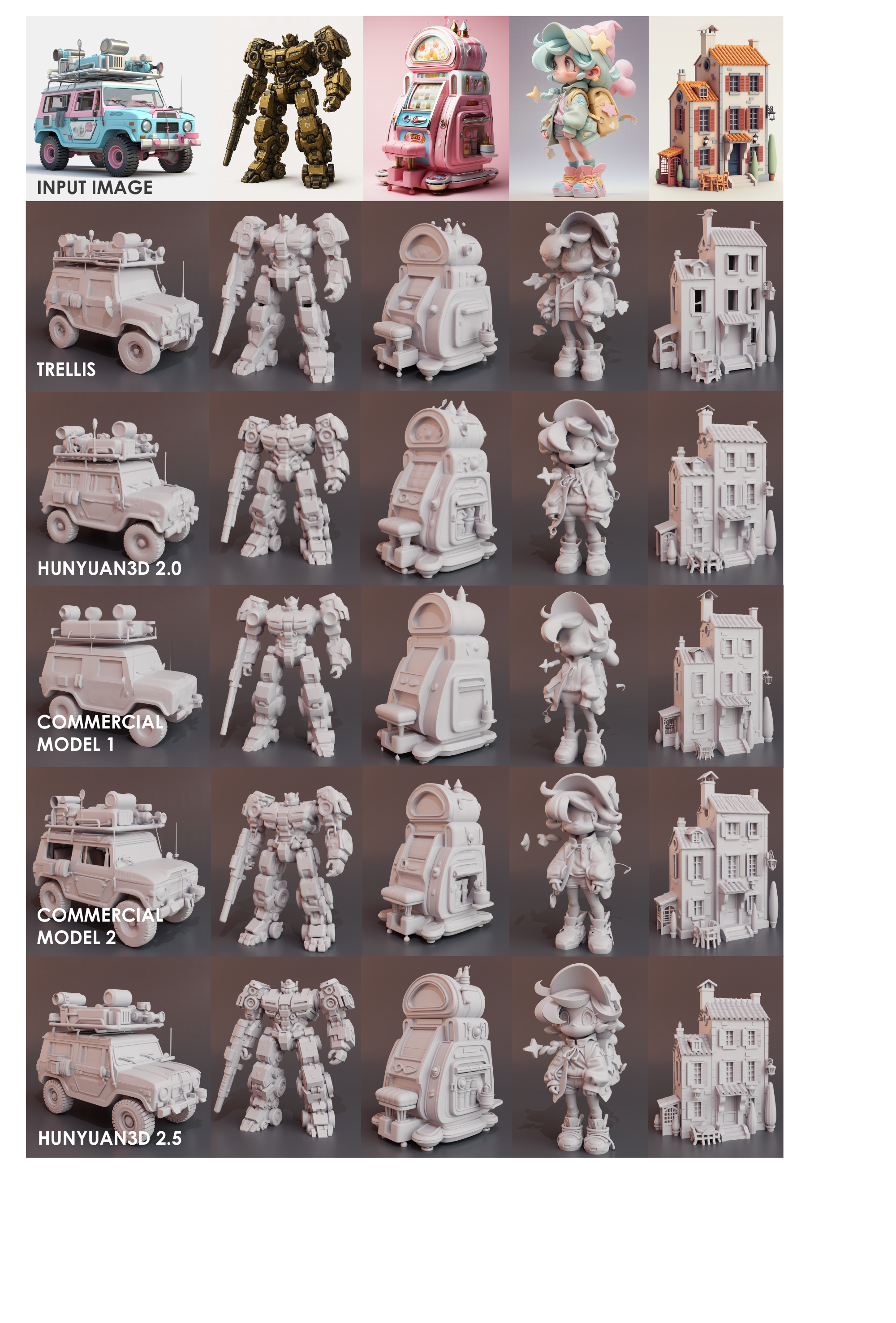}
    \caption{Visual comparison of different methods in terms of shape generation.}
    \label{fig:shape_results}
\end{figure}

\textbf{Comparison.} We show the numerical comparison in table.\ref{tab:shapegen} and visual comparison in \figref{fig:shape_results}. It can be observed that our method achieves the best image-shape and text-shape similarities in terms of ULIP-T and Uni3D-T and Uni3D-I. Nevertheless, we noted that these metrics could not fully reflect the model capabilities. As shown in \figref{fig:shape_results}, our model actually perform much better than all other open-sourced and commerical models.

\begin{figure}[H]
    \centering
    \includegraphics[width=1\textwidth]{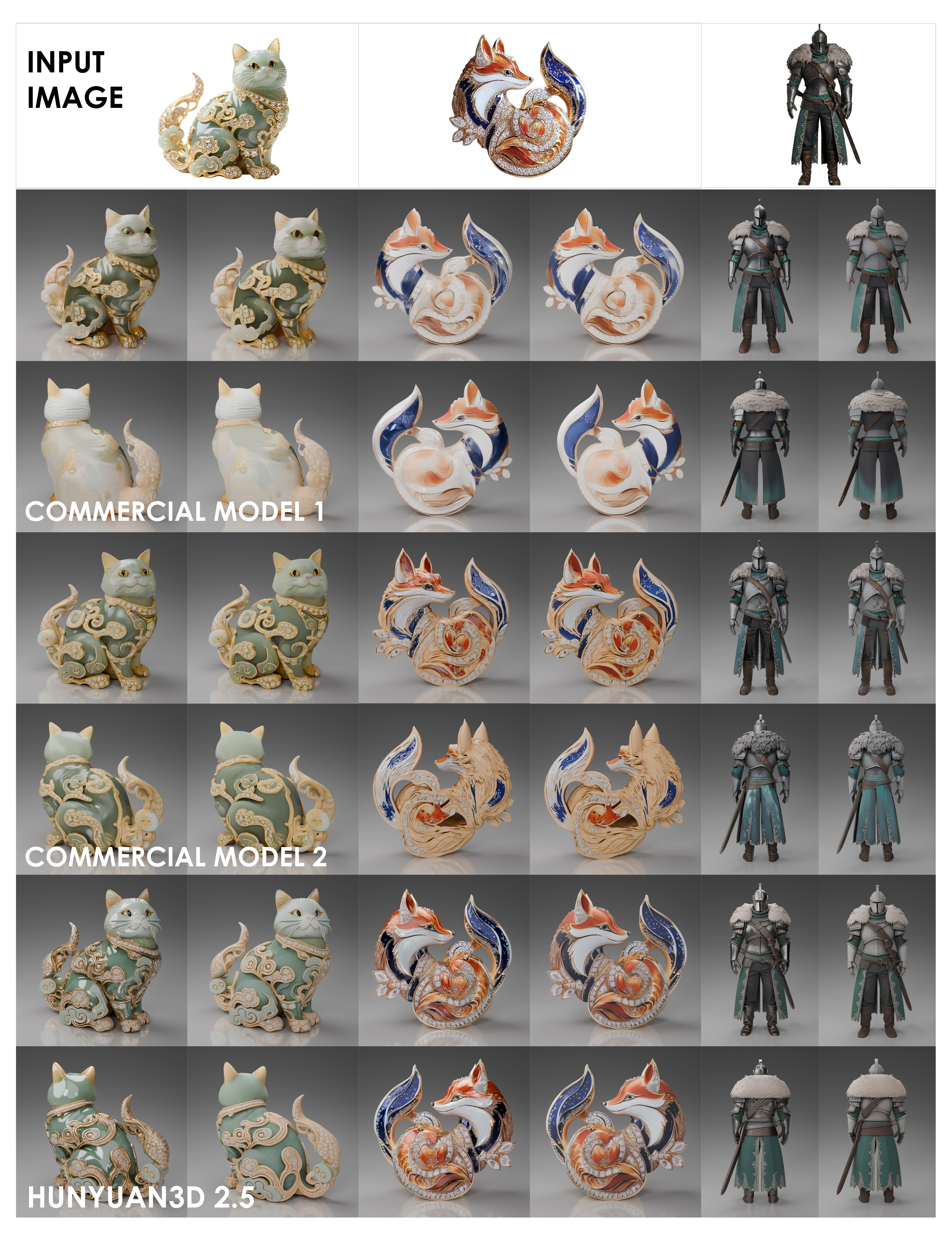}
    \caption{Visual comparison of different methods in terms of texture generation. We compared the front and back of models generated by different methods, as well as the effects of the corresponding complete material maps and albedo maps.}
    \label{fig:cmp_texture}
\end{figure}

\begin{table}[t]
\centering
\small
\caption{Numerical comparisons of different shape generation models  on ULIP-T/I, Uni3D-T/I.}
\begin{tabular}{rccccccc}
\toprule
                        & \textbf{ULIP-T($\uparrow$)} & \textbf{ULIP-I($\uparrow$)} & \textbf{Uni3D-T($\uparrow$)} & \textbf{Uni3D-I($\uparrow$)} \\ \midrule
Michelangelo~\citep{zhao2024michelangelo} & 0.0752 & 0.1152 & 0.2133 & 0.2611 \\
Craftsman 1.5~\citep{li2024craftsman}    & 0.0745 & 0.1296 & 0.2375 & 0.2987 \\
Trellis~\citep{xiang2024structured}      & 0.0769 & 0.1267 & 0.2496 & 0.3116 \\
Commercial Model 1 & 0.0741 & \textbf{0.1308} & 0.2464 & 0.3106 \\
Commercial Model 2 & 0.0746 & 0.1284 & {0.2516} & {0.3131} \\
Hunyuan3D 2.0~\citep{zhao2025hunyuan3d} & \underline{0.0771} & {0.1303} & \underline{0.2519} & \underline{0.3151} \\ 
\midrule
Hunyuan3D 2.5 & \textbf{0.07853} & \underline{0.1306}  & \textbf{0.2542}   & \textbf{0.3151}  \\ 
\bottomrule
\end{tabular}
\label{tab:shapegen}
\end{table}

\subsection{Texture Generation}
\textbf{Competing Methods.} 
We perform quantitive comparison with text- and image-conditioned methods, including
 Text2Tex \cite{chen2023text2tex}, Paint3D \cite{zeng2024paint3d}, Paint-it \cite{youwang2024paint}, SyncMVD \cite{liu2024text}, and TexGen \cite{yu2024texgen}. Furthermore, we show qualitive comparison on closed-source commercial models.

\textbf{Metrics.}
We use Fréchet Inception Distance (FID), CLIP-based FID (CLIP-FID), and Learned Perceptual Image Patch Similarity (LPIPS) to measure the similarity between the generated textures and the ground truth. CLIP Maximum-Mean Discrepancy (CMMD) is used to assess the diversity and richness of the generated texture details. And CLIP-Image Similarity (CLIP-I) is employed to evaluate how well the generated textures semantically align with the input images (for image prompt methods).

\textbf{Comparison.} We show the numerical comparison in table.\ref{tab: comparisons} and visual comparison in \figref{fig:cmp_texture}, For an intuitive comparison, we directly show the end-to-end results. 
It can be observed that for PBR material generation, competing models struggle to accurately estimate the correct MR (metallic and roughness) values, and face challenges in decoupling the inherent illumination effects in the input images for the albedo component.

\begin{table*}[t]
\centering
\caption{Quantitative comparison with state-of-the-art methods. We compare with two classes of methods, one conditioned on text only, and the other one based on image. Our method achieves the best performance compared with both classes.}
\setlength{\tabcolsep}{5pt}\small
\begin{tabular}{ccccccc}
\toprule 
Method     & {CLIP-FID$\downarrow$} & {FID$\downarrow$} & {CMMD$\downarrow$} & {CLIP-I$\uparrow$} & {LPIPS$\downarrow$}  \\ \midrule
Text2Tex \cite{chen2023text2tex} \textcolor{blue}{\textsubscript{ICCV'23}}  & 31.83    & 187.7      & 2.738 & -      & 0.1448 \\
SyncMVD \cite{liu2024text} \textcolor{blue}{\textsubscript{SIGGRAPH Asia'24}}   & 29.93    & 189.2      & 2.584 & -      & 0.1411 \\
Paint-it \cite{youwang2024paint} \textcolor{blue}{\textsubscript{CVPR'24}} & 33.54    & 179.1      & 2.629 & -      & 0.1538       \\
\hline
Paint3D \cite{zeng2024paint3d} \textcolor{blue}{\textsubscript{CVPR'24}} & 26.86    & 176.9      & 2.400 & 0.8871      & 0.1261       \\
TexGen \cite{yu2024texgen} \textcolor{blue}{\textsubscript{TOG'24}}  & 28.23    & 178.6      & 2.447 & 0.8818 & 0.1331       \\ 
Ours     & \textbf{23.97}    & \textbf{165.8}      & \textbf{2.064} & \textbf{0.9281} & \textbf{0.1231}       \\ \bottomrule
\end{tabular}
\label{tab: comparisons}
\end{table*}

\begin{figure}[h]
    \centering
    \includegraphics[width=0.95\textwidth]{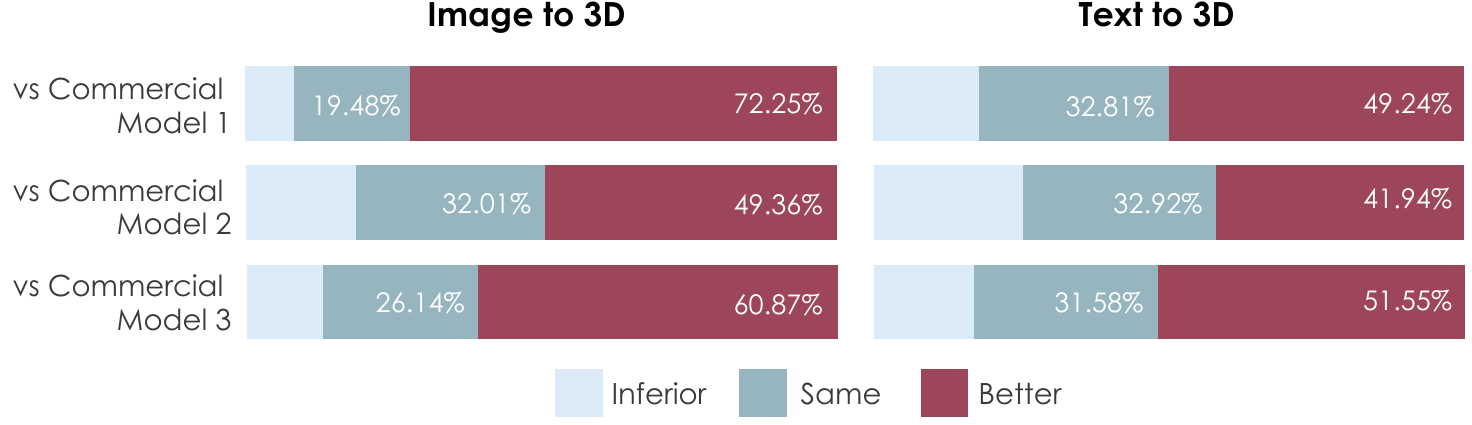}
    \caption{User study against three latest commerical models in terms of end-to-end textured results.}
    \label{fig:user_study}
\end{figure}

\textbf{User Study.} We also conducted a user study to evaluate human preferences for generated textured models using different methods. In this study, each participant was asked to rank each method for each sample in the testset. The testset included a diverse range of real-world images from various categories. As shown in \figref{fig:user_study}, we compared our method with three different commercial models. The results clearly demonstrate that our method significantly outperforms the others. For instance, in the image-to-3D task, our method achieved a 72\% win rate, which is 9 times higher than that of Commercial Model 1.

\section{Related Work}

\paragraph{Shape Generation.} 3D shape generation has advanced rapidly in recent years. Early works~\citep{wu2016learning,sanghi2022clip,yan2022shapeformer,yin2023shapegpt} based on different generative models~\citep{kingma2013auto,goodfellow2014generative,papamakarios2021normalizing} demonstrated the preliminary potential for generating specific categories of shapes. With the rise of diffusion models~\citep{rombach2022high, ho2020denoising}, 3D shape generation methods based on score distillation~\citep{poole2022dreamfusion} have been introduced, enabling text-to-3D generation by leveraging text-to-image models. Feedforward methods such as LRM~\citep{hong2023lrm}, Hunyuan3D 1.0~\citep{yang2024hunyuan3d} and LGM~\citep{tang2024lgm} represent another line of research focused on generating 3D assets in a single step. Recently, native 3D diffusion models have significantly improved generation quality by utilizing 3D data. Notable works in this area include Michelangelo~\citep{zhao2024michelangelo}, CLAY~\citep{zhang2024clay}, Trellis~\citep{xiang2024structured}, Hunyuan3D 2.0~\cite{zhao2025hunyuan3d}, and TripoSG~\citep{li2025triposg}, among others. Although multi-step sampling is required, native 3D diffusion models based on vecset~\cite{zhang20233dshape2vecset} can be accelerated via FlashVDM~\citep{lai2025flashvdm}, achieving speeds that surpass even feedforward methods. On the other hand, autoregressive models, such as MeshGPT~\citep{siddiqui2024meshgpt} BPT~\citep{weng2024scaling}, and Meshtron~\citep{hao2024meshtron} have become popular for mesh generation with human-like topology. 

\paragraph{Texture Synthesis.}
Multiview diffusion~\citep{zhao2025hunyuan3d,huang2024mvadapter,vainer2024jointlygeneratingmultiviewconsistent, li2024era3d, tang2025mvdiffusion++, long2024wonder3d, wang2023imagedream, shimvdream, shi2023zero123++, liu2023zero}, which primarily introduce cross-view attention mechanisms to modeling multi-view latent interactions, has opened up new avenues for addressing the global consistency issue of 3D textures. Zero123++~\citep{shi2023zero123++} spatially concatenates multi-view images and utilizes the self-attention to build up cross-view interaction. Other works inject view constraints into the attention block using diverse attention masks~\citep{Tang2023mvdiffusion, huang2024mvadapter, li2024era3d}. For PBR material generation, existing methods mainly include three categories: Generation-based approaches \cite{vainer2024collaborative, sartor2023matfusion, vecchio2024matfuse, chen2024intrinsicanything, zeng2024rgb} leverage diffusion models to learn material priors and recover PBR properties through physical rendering; retrieval-based techniques \cite{zhang2024mapa, fang2024make} adapt pre-built material graphs from libraries to ensure visual consistency and editability; optimization-based methods \cite{chen2023fantasia3d, zhang2024dreammat, wu2023hyperdreamer, xu2023matlaber, yeh2024texturedreamer, youwang2024paint, liu2024unidream} first generate initial textures and then refine them through techniques like Score-Distillation Sampling \cite{poole2022dreamfusion}.

\section{Conclusion}

In this work, we presented \shortname, an advanced suite of 3D diffusion models for generating high-quality, detailed 3D assets. By introducing a new shape foundation model and extending texture generation with physical-based rendering (PBR), \shortname achieves remarkable improvements in both shape fidelity and texture realism. Extensive evaluations show that \shortname outperforms current state-of-the-art models in terms of shape detail, surface smoothness, and texture consistency. This work marks a significant advancement in the 3D generation field, providing a powerful tool for creating realistic and detailed 3D assets across various industries.

\clearpage
\section{Contributors}

\textbf{Project Sponsors:} 

Jie Jiang, Linus 

\textbf{Project Leaders:} 

Chunchao Guo, Jingwei Huang, Zeqiang Lai

\textbf{Core Contributors:} 

\begin{itemize}[leftmargin=*]
    \item \textit{Shape Generation}: Zeqiang Lai, Yunfei Zhao, Jingwei Huang, Haolin Liu, Zibo Zhao, Qingxiang Lin, Huiwen Shi, Xianghui Yang
    \item \textit{Texture Generation}: Mingxin Yang, Shuhui Yang, Yifei Feng, Sheng Zhang, Xin Huang
\end{itemize}

\textbf{Contributors\footnote{Alphabetical order.}:} 

Di Luo, Fan Yang, Fang Yang, Lifu Wang, Sicong Liu, Yixuan Tang, Yulin Cai, Zebin He, Tian Liu, Yuhong Liu

\bibliography{colm2024_conference}
\bibliographystyle{colm2024_conference}

\end{document}